\documentclass{uai2022} 

\pdfoutput=1
\usepackage[american]{babel}

\usepackage{natbib} 
\usepackage{mathtools} 
\usepackage[table, dvipsnames]{xcolor}
\usepackage{booktabs, multirow} 
\usepackage{tikz} 
\usepackage{amssymb}
\usepackage{amsthm}
\usepackage{algorithm, algpseudocode}

\algnewcommand{\Initialize}[1]{\State \textbf{Initialize: \newline}}
\algnewcommand{\LineComment}[1]{\State  \(\triangleright\) #1 \hfill~}
\algnewcommand{\IIf}[1]{\State\algorithmicif\ #1\ \algorithmicthen\ }
\algnewcommand{\IElse}{\unskip\ \algorithmicelse\ }
\algnewcommand{\EndIIf}{\unskip}
\algnewcommand{\pElse}[1]{\State \algorithmicelse\ #1}

\algnewcommand{\FFor}[1]{\State\algorithmicfor\ #1 \algorithmicdo }
\algnewcommand{\EndFFor}{\unskip}

\theoremstyle{definition}

\newtheorem{proposition}{Proposition}
\newtheorem{theorem}{Theorem}

\DeclareMathOperator*{\argmax}{arg\,max}

\newcounter{step}[section]
\newcommand\step{\refstepcounter{step}\par \textit{\underline{ Step \thestep}:~}}

\usepackage{tikz}
\usepackage{pgfplots}
\usetikzlibrary{automata, positioning, shapes.geometric, arrows, decorations, fit, backgrounds}
\usetikzlibrary{arrows.meta}
\usepgfplotslibrary{statistics}
\usetikzlibrary{backgrounds}
\pgfplotsset{compat=1.17}
\tikzset{reverseclip/.style={insert path={(-100cm,100cm) rectangle (100cm,-100cm)}}}

\makeatletter
\let\MYcaption\@makecaption
\makeatother
\usepackage[font=footnotesize]{subcaption}
\makeatletter
\let\@makecaption\MYcaption
\makeatother

\makeatletter
\newcommand*{\rom}[1]{\expandafter\@slowromancap\romannumeral #1@}
\makeatother

\title{MPE inference using an Incremental Build-Infer-Approximate Paradigm}

\author[1]{\href{mailto:<ee13s064@ee.iitm.ac.in>?Subject=Your UAI 2022 paper}{Shivani Bathla}{}}
\author[1]{Vinita Vasudevan}
\affil[1]{%
    Electrical Department\\
    IIT Madras\\
    Chennai, India
}
  
\begin{document}
    
\maketitle

\begin{abstract}
 Exact inference of the most probable explanation (MPE) in Bayesian networks is known to be NP-complete. In this paper, we propose an algorithm for approximate MPE inference that is based on the incremental build-infer-approximate (IBIA) framework. We use this framework to obtain an ordered set of partitions of the Bayesian network and the corresponding max-calibrated clique trees. We show that the maximum belief in the last partition gives an estimate of the probability of the MPE assignment. We propose an iterative algorithm for decoding, in which the subset of variables for which an assignment is obtained is guaranteed to increase in every iteration. There are no issues of convergence, and we do not perform a search for solutions. Even though it is a single shot algorithm, we obtain valid assignments in 100 out of the 117 benchmarks used for testing. The accuracy of our solution is comparable to a branch and bound search in majority of the benchmarks, with competitive run times. 
\end{abstract}

\section{Introduction}\label{sec:intro}
Bayesian Networks (BN) are directed graphical models that provide a compact representation of joint probability distribution over a set of random variables. An important task in bayesian inference is to find the \textit{Most Probable Explanation (MPE)} given a set of evidence variables. It involves finding a complete assignment for all non-evidence variables that gives the maximum probability. The probability of the MPE assignment is same as the \textit{max-marginal} obtained after maximizing the joint distribution over all non-evidence variables. Exact methods include variable elimination and max-product belief propagation on join trees followed by a decoding step to get the MPE assignment  and integer linear programming (ILP) ~\citep{Koller2009}.
However,  MPE inference is known to be an NP-complete problem~\citep{Shimony1990, Park2004} and exact methods are not viable in most cases.

One approach to get the MPE assignment is to use search based algorithms that traverse the model's search space. A commonly used framework for graphical models is the AND/OR search tree~\citep{Dechter2007}. Several search algorithms have been used to find MPE estimates from these models including depth-first Branch and Bound (AOBB)~\citep{Marinescu2005,Marinescu2014,Marinescu2018}, best-first (AOBF)~\citep{Marinescu2007,Marinescu2020}, Breadth-Rotating AND/OR Branch and Bound (BRAOBB)~\citep{Otten2012}. These methods use upper bounds of max-marginals computed using different variants of Mini-Bucket Elimination (MBE) to guide the search. Weighted Mini-Bucket elimination improves the accuracy of the bound of traditional MBE using the Holder's inequality~\citep{Liu2011}. Another technique to tighten the resulting bound is to use cost-shifting schemes. Two such algorithms are Mini-Bucket Elimination with Moment Matching (MBE-MM)~\citep{Flerova2011} and Join Graph Linear Programming (JGLP)~\citep{Ihler2012}. While MBE-MM uses a single pass of cost-shifting locally within the  mini-buckets, JGLP performs cost-shifting updates to the entire mini-bucket join graph iteratively. 

Another class of methods are based on LP relaxations.  LP relaxations based on message passing include tree re-weighted max-product~\citep{Wainwright2005,Kolmogorov2005,Werner2007,Jojic2010,Hazan2010} and max-product linear programming~\citep{Globerson2007,Sontag2008}. The other class of methods are the dual decomposition solvers~\citep{Komodakis2007,Sontag2011,Sontag2012,Ping2015,Martins2015}. The connection  between the solutions obtained using two methods is explored in \citet{Bauer2019}


In this work, we propose a different approach for MPE inference. Our method is based on an Incremental Build-Infer-Approximate  paradigm (IBIA) that returns a set of bounded clique size partitions of the BN along with the calibrated forest of of clique trees (CTFs) corresponding to each partition \citep{JAIR}. Instead of the sum-product calibration used in \citet{JAIR}, we use a max-product calibration along with suitable modifications to the Approximate step. Based on this, we propose a single search algorithm for MPE inference with the following features.
\begin{enumerate}
\item We show that the max-marginal can be estimated  by finding the maximum belief from any clique in the max-calibrated CT of the last partition. Based on experiments, we find that our algorithm gives very good estimates with an average error of $\pm 0.27$ in the log probability over 79 benchmarks for which the exact solution is available.
\item We propose a single shot algorithm for decoding the MPE assignment. It is an iterative method in which the subset of variables for which an assignment is obtained is guaranteed to increase in every iteration. There are no issues of convergence. Even though it is a single shot algorithm, we were able to get a non-zero probability assignment with accuracy comparable to AOBB in majority of the testcases.
 \item Our code is written in Python, with Numpy and NetworkX libraries. The runtimes of our algorithm are very competitive and better in some cases, than the compiled C++ code for AOBB.
\end{enumerate}


\section{Background}\label{sec:background}
A Bayesian network  is a directed acyclic graph (DAG), $\mathcal{G}$. Its nodes represent random variables $\mathcal{Y} = \{ X_1, X_2, \cdots X_n\}$ with associated domains $D = \{D_1,D_2, \cdots D_n\}$. It has directed edges from a subset of nodes $\textrm{Pa}_{X_i} = \{X_k, k \neq i\}$ to $X_i$, representing a conditional probability distribution (CPD) $\phi_i = P\{X_i|\textrm{Pa}_{X_i}\}$. The BN represents the joint probability distribution (JPD) of $\mathcal{Y}$, given by $P(\mathcal{Y}) = \prod_{i=1}^n \phi_i$.
Let $E$ denote the set of instantiated or evidence variables, with $\mathcal{Y} = \{\mathcal{X}, E\}$. The task is to find the most probable explanation (MPE), defined as
\begin{align*}
     \textrm{MPE}=\argmax\limits_{\mathcal{X}} P(\mathcal{X},E=e)
 \end{align*}

In low treewidth networks, MPE can be obtained using the clique tree (CT) representation, which is a derived undirected graphical model obtained after moralization and chordal completion of the BN.  The CT, which is also called the join tree or junction tree,  is a hypertree with nodes $\{C_1,C_2,\cdots, C_n\}$ that are the set of maximal cliques in the chordal graph.  An edge between $C_i$ and $C_j$ is associated with a sepset $S_{i,j} = C_i \cap C_j$. Each factor $\phi_i$ is associated with a single clique $C_i$ such that $\textrm{Scope}(\phi_i) \subseteq \mathcal{C}_i$. 
The task of computing MPE is split into two parts. First, the max-product belief propagation (BP) algorithm~\citep[Chapter 10 of][]{Koller2009} is used to compute the maximum marginal probability, $MaxMarg$, defined as
\begin{align*}
MaxMarg &= \max\limits_{\mathcal{X}} P(\mathcal{X},E=e) 
\end{align*}
 This algorithm has two rounds of message passing along the edges of the CT, an upward pass (from the leaf nodes to the root node) and a downward pass (from the root node to the leaves).
 The CT obtained as a result of this algorithm is \textit{max-calibrated},  which is defined as follows.
Let $\beta(C_i)$, $\beta(C_j)$ denote the beliefs associated with adjacent cliques $C_i$ and $C_j$ and $\mu_{ij}$, the belief of the corresponding $S_{ij}$. The CT is said to be max-calibrated if all pairs of adjacent cliques satisfy the following. 
   \begin{equation*}
   \max\limits_{C_i\setminus S_{i,j}}\beta(C_i) =   \max\limits_{C_j\setminus S_{i,j}}\beta(C_j) = \mu(S_{i,j})
   \end{equation*}
The $MaxMarg$ can be simply obtained by finding the maximum clique belief in any clique $C$ in the max-calibrated CT since,
\begin{align*}
  MaxMarg = 
  \max\limits_{C}\max\limits_{\mathcal{X}\setminus C} P(\mathcal{X},e)
  = \max\limits_{C} \beta(C)
\end{align*}
Since a network could have disjoint DAGs, we use the term \textit{Clique Tree Forest} (CTF) to denote the collection of corresponding CTs. The max-marginal for a BN with multiple DAGs is the product of max-marginals of all CTs $\in CTF$. 

Once $MaxMarg$ is computed, the decode step finds an MPE assignment using the max-calibrated CTF. Algorithm~\ref{alg:traceback} has the traceback procedure. Set $S_u$ tracks the set of unassigned variables. For each CT in the CTF, any clique can be chosen as the root node ($C_{root}$). 
To start with all variables in the root clique are unassigned. Since the global MPE assignment also satisfies the local optimality property~\citep{Koller2009}, a partial assignment can be obtained by finding the argmax of the clique belief. The overall MPE assignment is updated and all assigned variables are removed from set $S_{u}$. All other cliques are iterated through in a pre-order traversal. For any clique $C$, the state of the variables on the sep-sets with the preceding cliques are known. After reducing belief $\beta(C)$ over the known variable states,  we find an assignment of the unassigned variables that maximizes the reduced clique belief.
\begin{algorithm}
    \scriptsize
	\caption{Traceback~($CTF$)}	\label{alg:traceback}
	\begin{algorithmic}[1]
	 \Require $CTF$: Max-calibrated CTF
	 \State \textbf{Initialize:\newline} \indent MPE$=~<>$ \Comment{{\color{teal!70} \scriptsize Hash table $<var: var.assignment>$}}\newline
	 \indent$S_u \gets$ Variables $\in CTF$ 	 \Comment{{\color{teal!70} Set of unassigned variables}}
	 \For{$CT\in CTF$}
	    \State $C_{root}\gets$ Choose any clique in CT
	    \For{$C \in CT.Preorder\_Traversal(C_{root})$}
	        \State $S_{u_c}\gets C\cap S_u$ \Comment{{\color{teal!70} Set of unassigned variables in clique C}}
	    \If{$S_{u_c} \neq \varnothing $}
	        \State $\beta(C)$ $\gets$ Reduce belief $\beta(C)$ over known states $\in$ MPE
	        \State A $\gets$ argmax $\beta(C)$
	        \State MPE[v] = A[v] $\forall v\in S_{u_c}$ \Comment{{\color{teal!70} Update MPE assignment}}
	        \State $S_u$.remove($S_{u_c}$)
	     \EndIf
	     \EndFor
	 \EndFor
    \State \Return MPE
	\end{algorithmic}
\end{algorithm}

\subsection{IBIA framework}
\begin{figure*}
\includegraphics[width=\textwidth]{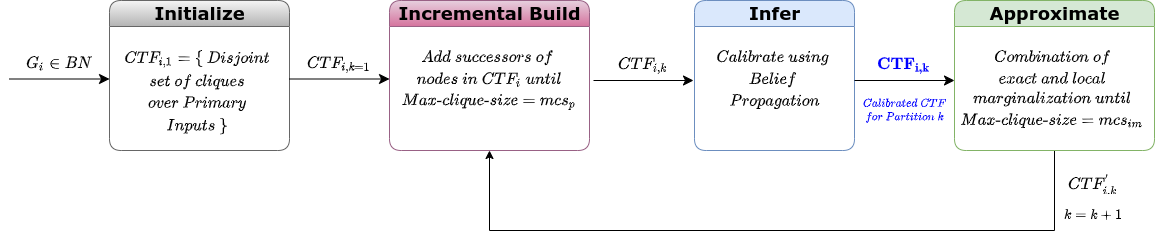}
\caption{Bayesian inference using the IBIA framework}
\label{fig:ibiaFramework}
\end{figure*}
The IBIA framework proposed in~\citet*{Bathla2021,JAIR}  presents a three-step approach to perform inference on a BN. The overall approach is illustrated in Figure~\ref{fig:ibiaFramework}. Each DAG $G_i\in BN$ is built, inferred and approximated separately. 
IBIA starts with a clique forest over the nodes that have no parents (referred to as \textit{primary inputs (PI)}). An incremental approach is used to add as many nodes as possible until a preset bound ($mcs_p$) on the clique size is reached. In this work, we define \textit{clique-size} for a given clique $C_i$ as,
  \begin{equation*}\label{eqn:cs}
    cs_i = \log_2~(\prod\limits_{\forall~v~\in~ C_i} |D_v|)
  \end{equation*}
  where $|D_v|$ is the cardinality or  the number of states in the domain of the variable $v$. 
When the clique-size bound $mcs_p$ is reached, the CTF is calibrated using the standard join tree message passing algorithm \citep{Koller2009}.
The CTF contains two sets of variables - variables that are parents to nodes that have not yet been added (referred to as {\it interface nodes}) and other variables ({\it non-interface nodes}). 
A combination of exact and approximate marginalization is used to lower clique sizes to a preset bound $mcs_{im}$, thus allowing for addition of new nodes.
This approximate CTF is used as the starting point for incremental construction of the CT for the next partition. This process is continued until all the nodes in the BN have been added to some CTF. 

 The algorithm returns an ordered set of partitions $\{R_1,R_2, \cdots R_P\}$ and the corresponding calibrated CTFs for each DAG $\in BN$. 
The authors show that the CTFs obtained after the build and approximate stage contain valid CTs. Also, the approximation algorithm maintains consistency of within-clique beliefs. A trade-off between runtime and accuracy can be achieved using the input parameters, $mcs_p$ and $mcs_{im}$.

\section{Framework for MPE queries}
The IBIA framework has been used to infer the partition function and the prior and posterior marginal probabilities of non-evidence variables in \cite{JAIR}. We have modified this framework  for MPE queries. We use an iterative procedure that has the build, infer and approximate steps and a partial decode step that gives an MPE assignment for a subset of variables. In each iteration, the subset for which an assignment is obtained is extended. The iteration is continued until either all variables are assigned or an assignment is not possible. In the following sections, each of the steps are explained in more detail.
\step{\textit{Incremental Build:}}
The CTF can be built incrementally using either of the algorithms detailed in \citet*{Flores2002,JAIR}.
We use the approach described in~\citet{JAIR} since it works directly with the clique tree model and eliminates the need for computation of any other intermediate representation. Typically, it chooses a smaller graph chosen for re-triangulation than~\citet{Flores2002}.
\step{\textit{Inference:}}
We use the two-pass Max-product Belief Propagation algorithm to calibrate the partition CTF. After calibration, all cliques in a CT will have the same maximum probability, and the beliefs of adjacent clique beliefs agree over the common variables~\citep{Koller2009}.
\step{\textit{Approximation:}}
\begin{figure}
    \centering
    \begin{subfigure}[b]{0.23\textwidth}
     \centering
    \includegraphics[scale=0.25]{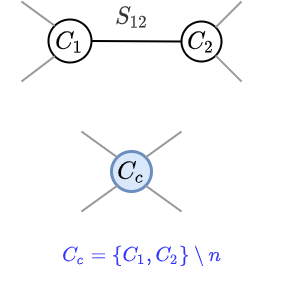}
    \caption{ Exact max-marginalization}
    \label{fig:exact}
    \end{subfigure}\hfill
    \begin{subfigure}[b]{0.25\textwidth}
    \centering
    \includegraphics[scale=0.25]{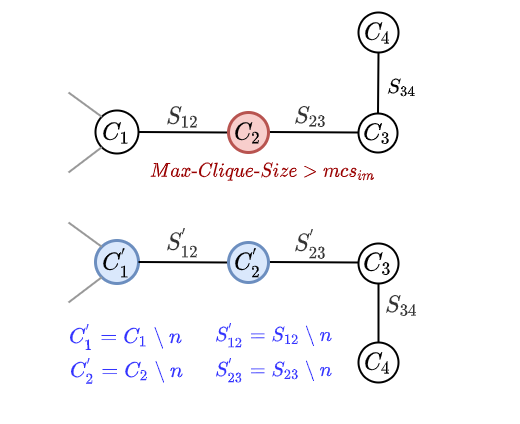}
    \caption{Local max-marginalization}
    \label{fig:local}
    \end{subfigure}
    \caption{Illustration of exact and local max-marginalization of variable $n$ present in cliques $C_1, C_2$ in first figure and cliques $C_1,C_2,C_3,C_4$ in the second.}
\end{figure}
As discussed, the nodes present in the CTF corresponding to a partition can divided into two sets: interface (IV) and non-interface nodes (NIV). The interface nodes are parents of the variables that have not yet been added. Therefore, these are required for building the next partition. We adopt a similar approximation methodology as suggested in~\citet{JAIR} to reduce the clique sizes. We also denote the calibrated tree obtained after BP as $CTF_{in}$ and the approximated CTF as $CTF_a$. The minimal subgraph of $CTF_{in}$ that connects the interface variables forms the initial $CTF_a$. This graph may also contain some non-interface variables. We use a combination of exact and local max-marginalization to reduce the clique sizes to a preset bound $mcs_{im}$. We now discuss both these steps in detail. 
\begin{enumerate}
    \item \textit{Exact max-marginalization:} If a non-interface variable $n$ is present in a single clique, it is simply removed from the clique scope and the belief is modified as: $\beta(C^{'}) = \max\limits_{n.states} \beta(C)$ where, $C'=C\setminus n$.
    On the other hand, if $n$ is present in multiple cliques, the subtree over all containing cliques ($ST_n$) is collapsed into a single clique ($C_c$). Figure~\ref{fig:exact} shows an example. Cliques $C_1,C_2$ containing $n$ are collapsed to form clique $C_c$.  Since this step could potentially result in a large clique, this is only performed if the size of the resultant clique $\leq mcs_{im}$. 
    The belief of collapsed clique is computed as shown below.
    \begin{equation*}
  \beta(C_c) = \max_{n.states} \left ( \frac{\prod_{C\in {ST_n}} \beta(C)}{\prod_{SP \in {ST_n}}\mu(SP)} \right)
\end{equation*}
where,  $SP$ denotes the sepsets in $ST_n$. \\
This step preserves the joint beliefs i.e. variables in $CTF_a$ have the same belief as in $CTF_{in}$.
\item \textit{Local max-marginalization:} In this step, variables present in cliques with size $>mcs_{im}$ are max-marginalized from individual cliques while ensuring (a) All CTs $\in CTF_a$ are valid CTs. (b) A connected CT $\in CTF_{in}$ remains connected in $CTF_a$. 

Let, $n$ be a variable present in large cliques. The subgraph of $CTF_{in}$ over all cliques containing $n$ is a connected tree since the input CTF is valid. To ensure that the \textit{Running Intersection Property (RIP)}~\citep{Koller2009} is satisfied after approximation, $n$ is retained in a connected subtree containing smaller cliques with sizes $\leq mcs_{im}$. It is individually max-marginalized out from all other cliques. Figure~\ref{fig:local} shows an example where variable $n$ is removed from cliques $C_1, C_2$ and retained in cliques $C_3,C_4$. Local marginalization of $n$ from cliques $C_i$ and $C_j$, and the corresponding sepset $S_{i,j}$ results in cliques $C_i' = C_i\setminus n$ and $C_j'=C_j\setminus n$ with sepset $S_{i,j}' = S_{i,j}\setminus n$. The corresponding beliefs are $\beta(C_i') = \max\limits_{n.states}\beta(C_i)$, $\beta(C_j')  = \max\limits_{n.states}\beta(C_j)$, $\mu(S_{i,j}') =  \max\limits_{n.states}\mu(S_{i,j})$.
While updating the beliefs in this manner preserves the joint beliefs for variables present within the clique, the joint beliefs of variables present in disjoint cliques are approximated. 
\end{enumerate}
Once the clique sizes are reduced, we re-assign the clique factors based on calibrated clique and sep-set beliefs. 

\begin{proposition} \label{pr:approx1}
  If all CTs in $CTF_{in}$ are valid and max-calibrated, then all CTs in $CTF_a$ are also valid and max-calibrated.
\end{proposition}
\begin{proposition}\label{pr3}
Approximation algorithm preserves the maximum belief and the within-clique beliefs of all cliques in $CTF_a$.
\end{proposition}
We discuss the proofs for Propositions~\ref{pr:approx1}, ~\ref{pr3} in Section~\ref{sec:sup}.


Algorithm~\ref{alg:ApproximateCTF} shows the overall algorithm. At all points in the method, we remove any non-maximal cliques if generated and re-connect their neighbors. We start by identifying the set of interface and non-interface variables present in $CTF_{in}$. We initialize $CTF_a$ as the minimal subgraph of $CTF_{in}$ that connects the interface variables.   We first attempt to reduce clique sizes using exact max-marginalization of non-interface variables. NIVs present in a single clique are removed, and the corresponding clique beliefs are max-marginalized. If any resultant clique is non-maximal, it is removed, and the neighbors are re-connected. Wherever possible within the clique size constraints, NIVs present in multiple cliques are max-marginalized after collapsing the cliques (lines 4-7).

Next, we find the set of large-sized cliques ($L_c$) with size $> mcs_{im}$. All variables in these cliques are grouped into NIVs and IVs and stored in $N$. We first try to locally max-marginalize non-interface variables. This is done as follows. First, both sets of variables are individually sorted in the increasing order of the number of containing cliques. Starting with NIVs, we pop a variable ($n$) from this list and identify the subgraph of $CTF_a$ over cliques containing this variable ($ST_n$). Since the input CTF satisfies RIP, $ST_n$ is a connected tree. $ST_n$ can be further divided into disjoint subtrees over cliques with size $\leq mcs_{im}$. We choose the largest subtree ($ST_r$) and retain the variable $n$ in all cliques in this subtree (lines 13-15). Since we want a connected CT to remain connected, the variable is removed only if no sepset size becomes zero (lines 15-18).

If the clique sizes remain above $mcs_{im}$ after processing NIVs, we start with the IVs. If $n$ is an IV, it must be retained in atleast one clique. Therefore, we ignore interface variables that are only present in large cliques (lines 19-20). Otherwise, we max-marginalize all clique and sep-set beliefs associated with the cliques that are not present in $ST_r$. The list $L_c$ is updated by removing cliques where the size is now within the preset limit. This process is repeated until either clique list $L_c$ or the variable set $N$ becomes empty.   

Once clique-sizes are reduced, we reassign the clique factors (lines 26-27). For each clique tree in $CTF_a$, we pick a root node and perform a pre-order traversal from the root node to all other nodes in the CT. The factor associated with the root node is the corresponding clique belief. For all other nodes, the factor for an unvisited neighbor $C_j$ of clique $C_i$ is assigned as $\beta(C_j)=\frac{\beta(C_i)}{\mu(S_{ij})}$.
\begin{algorithm}[t]
	\scriptsize
	\caption{ApproximateCTF~($CTF, ~mcs_{im}$)}
	\label{alg:ApproximateCTF}	
	\begin{algorithmic}[1]
		\Require~$CTF_{in}$: Input clique tree forest \newline
		\indent$mcs_{im}$: Maximum clique size limit for interface $CTF_{in}$
		\Ensure$CTF_a$: Approximate $CTF_{in}$\newline
		\State $IV, NIV\gets$ Divide nets $\in CTF_{in}$ into interface and non-interface variables
		\State $CTF_a \gets$ Minimal subgraph of $CTF_{in}$ connecting $IV$
		\State $NIV\gets Nets\in CTF_a\setminus IV$ \Comment{{\color{teal!70}\scriptsize Updated set of Non-interface variables}}
		\LineComment{{\color{teal!70}Exact max-marginalization}} 
        \State Max-marginalize NIVs present in a single clique in $CTF_a$
		\State Remove resultant non-maximal cliques, reconnect neighbors
		\State Collapse cliques to max-marginalize NIVs in multiple cliques if $cs< mcs_{im}$
        \LineComment{{\color{teal!70}Local max-marginalization to trim cliques with size $> mcs_{im}$}}
		\State $L_c\gets$ List of cliques with size $> mcs_{im}$
		\State $N\gets \{NIV,IV\}$ \Comment{{\color{teal!70} Sorted lists of variables present in cliques in $L_c$}}
		\While{$L_c\neq \varnothing$ or $N\neq \varnothing$} 
		\State $n=N.$pop net present in least number of cliques
		\State $ST_n\gets$ Subgraph of $CTF_a$ over cliques containing $n$
		\State $ST_r\gets $ Choose connected subtree $\in ST_n$ s.t. max-clique-size $\leq mcs_{im}$
		\LineComment{{\color{teal!70}Check if CT will remain connected if $n$ is removed}}
		\State $L_m\gets$Cliques $\in ST_n ~\setminus~$ Cliques $\in ST_r$
		\State  $ms\gets$ Minimum sep-set size for cliques in $L_m$
		\IIf{$ms==1$} continue
		\LineComment{{\color{teal!70}Check if interface variables are retained in atleast one clique}}
		\IIf{$ST_r.isEmpty() ~\&\&~ n\in IV$} continue
		\State Locally max-marginalize beliefs corresponding to all cliques in $L_m$. 
		\State Remove resultant non-maximal cliques, reconnect neighbors
		\State $L_c$.remove($C$)  if $C$.size $\leq mcs_{im}$	 $\forall ~C\in L_m$  
		\EndWhile
		\LineComment{{\color{teal!70}Re-parameterize the joint distribution}}
		\State Re-assign clique factors in $CTF_a$ using clique and sep-set beliefs
	\State \Return $CTF_a$
	\end{algorithmic}
\end{algorithm}

\step{\textit{Decode:}}
The assignment of variables in the last partition is used as the partial assignment of MPE. This assignment is obtained from the last partition CT using Algorithm~\ref{alg:traceback}. 
\subsection{Inference of $MaxMarg$}
Let $CTF_i$ denote the CTF for the $i^{th}$ DAG in the BN. The $n_i$ partitions of the $i^{th}$ DAG are denoted $\{R_{i1}, R_{i2}, \hdots, R_{i{n_i}}\}$, with corresponding CTFs  $CTF_i = \{CTF_{i,1}, CTF_{i,2}, \hdots, CTF_{i,{n_i}}\}$. 
\begin{proposition}\label{pr4}
The max-belief of any clique in $CT\in CTF_{i,k}$ is the estimate of the maximum probability in the joint distribution over all variables added to $CT$ in partitions $R_{ij}, j\leq k$.
\end{proposition}
\begin{proof}

Let, $C$ be a clique in the calibrated $CT\in CTF_{i,1}$ and $\mathcal{X}_1$ be the set of variables in $CT$. Cliques in $CT$ are assigned CPDs $\{\phi_v, v\in\mathcal{X}_1 \}$ as factors. Therefore, the following holds true once $CT$ is calibrated.  
\begin{align*}
    \beta(C) &= \max\limits_{\mathcal{X}_1\setminus C} \prod\limits_{v\in\mathcal{X}_1} \phi_v
    = \max\limits_{\mathcal{X}_1\setminus C} P(\mathcal{X}_1)\\
    \implies  &\max\limits_{C}\beta(C) = \max\limits_{ \mathcal{X}_1}P(\mathcal{X}_1)
\end{align*}
Let  $CTF_{k,1}^{'}$ be the CTF obtained after approximating  $CTF_{k,1}$. While the within clique beliefs are preserved while local max-marginalization (by Proposition~\ref{pr3}), the joint beliefs of variables belonging to different cliques are approximated. Let, $\mathcal{X}_{12}$ be the set of variables in $CTF_{k,1}^{'}$ and $P'(\mathcal{X}_{12})$ denote the joint belief obtained from the approximate beliefs in $CTF_{k,1}^{'}$ as follows. 
Let, $Q(\mathcal{X}_1)$ denote the distribution such that 
\begin{align*}
    \max\limits_{\mathcal{X}_1\setminus \mathcal{X}_{12}} Q(\mathcal{X}_1) = P'(\mathcal{X}_{12})= \dfrac{\prod \beta^{'}}{\prod \mu^{'}}
\end{align*}

Let, $\mathcal{X}_2$, denote the set of new variables added to the corresponding $CT\in CTF_{i,2}$ in the second partition. Factors assigned to cliques in CT are the CPDs of the new nodes $\{\phi_v, ~\forall v \in \mathcal{X}_2\}$ and re-parameterized beliefs $\{\beta^{'}_1,\frac{\beta^{'}_2}{\mu^{'}_{12}}, \hdots\}$. Therefore, after calibration of CT, the following holds true for any clique $C\in CT$.
\begin{align*}
\beta(C) &= \max\limits_{\{\mathcal{X}_2, \mathcal{X}_{12}\}\setminus C} \prod\limits_{v\in \mathcal{X}_2} \phi_v \cdot \frac{\prod \beta^{'}}{\prod \mu^{'}}\\
&= \max\limits_{\{\mathcal{X}_2, \mathcal{X}_{12}\}\setminus C} P(\mathcal{X}_2~|~Par_{\mathcal{X}_2}) \cdot P'(\mathcal{X}_{12}) \\
&= \max\limits_{\{\mathcal{X}_2, \mathcal{X}_{12}\}\setminus C} P(\mathcal{X}_2~|~Par_{\mathcal{X}_2}) \cdot \max\limits_{\mathcal{X}_1\setminus \mathcal{X}_{12}}Q(\mathcal{X}_{1}) \\
&= \max\limits_{\{\mathcal{X}_2, \mathcal{X}_{1}\}\setminus C} P(\mathcal{X}_2~|~Par_{\mathcal{X}_2}) \cdot Q(\mathcal{X}_{1})
\end{align*}
Since variables in $\mathcal{X}_1$ are non-descendants of $\mathcal{X}_2$ in the BN, ${\mathcal{X}_2 \perp \mathcal{X}_1 ~|~ Par_{\mathcal{X}_2}}$. Therefore,
\begin{align*}
    P(\mathcal{X}_2~|~\mathcal{X}_1, Par_{\mathcal{X}_2}) &= P(\mathcal{X}_2~|~Par_{\mathcal{X}_2})\\
    P(\mathcal{X}_2~|~\mathcal{X}_1) &= P(\mathcal{X}_2~|~Par_{\mathcal{X}_2}) ~~\because~ Par_{\mathcal{X}_2}\subset \mathcal{X}_1
\end{align*}
Therefore, the clique belief can be written as,
\begin{align*}
\beta(C)
&= \max\limits_{\{\mathcal{X}_2, \mathcal{X}_{1}\}\setminus C} P(\mathcal{X}_2~|~\mathcal{X}_1) \cdot Q(\mathcal{X}_1)\\
&= \max\limits_{\{\mathcal{X}_2, \mathcal{X}_1\}\setminus C} P''(\mathcal{X}_2,\mathcal{X}_1)\\
\max\limits_{C}\beta(C) &= \max\limits_{\mathcal{X}_2, \mathcal{X}_1} P''(\mathcal{X}_2,\mathcal{X}_1)
\end{align*}
A similar procedure is repeated in each successive partition. Therefore, the max-belief in the calibrated CT in partition $R_{ik}$ approximates the maximum probability of the joint distribution over all variables added to it in partitions $R_{ij}, j\leq k$.
\end{proof}
\begin{theorem}\label{thm1}
The product of the max-belief of the CTs corresponding to the
last partition of all DAGs is the estimate of $MaxMarg$. 
\end{theorem}
\begin{proof}
For each DAG, $G_i$, the last partition CTF contains a single CT since we never break a connected CT. Using Proposition\ref{pr4}, the maximum probability ($MaxMarg_i$) for this CT is the estimate of the maximum probability of all variables in $G_i$. This value is exact when there is only one partition and may be approximate when there are multiple partitions. The product $\prod_i MaxMarg_i$ is the estimate of the overall max-marginal $MaxMarg$.
\end{proof}
Based on Theorem~\ref{thm1}, Algorithm~\ref{alg:findMaxMarg} computes the overall max-marginal.

\begin{algorithm}
    \scriptsize
	\caption{FindMaxMarg~($L_{CTF}$)}	\label{alg:findMaxMarg}
	\begin{algorithmic}[1]
	  \Require $L_{CTF}$: List of clique tree forest corresponding to all DAGs $\in$ BN 
	 \Ensure $MaxMarg$: Estimated max-marginal over non-evidence variables $\in$ BN
     \State $MaxMarg=1.0$
     \For{$L_i \in L_{CTF}$} \Comment{{\color{teal!70} $L_i$: List of partition CTFs for the i\textsuperscript{th} DAG}}
     		\State CT $\gets L_i[-1]$ \Comment{{\color{teal!70}  Calibrated CT for the last partition}}
            \State $C\gets$ Choose any clique in CT
           	\State $MaxMarg = MaxMarg \times \max\limits_{C} \beta(C)$   \Comment{{\color{teal!70} \scriptsize Multiply with max clique belief}}
     \EndFor
    \State \Return $MaxMarg$
	\end{algorithmic}
\end{algorithm}



\subsection{MPE Decoding and complexity}
\begin{algorithm}[t]
\caption{MPE inference procedure}\label{alg:inferMPE}
\scriptsize
    \begin{algorithmic}[1]
     \State Initialize: MPE $\gets \{\}$; $M = \{\textrm{MPE}, E\}$;
      \State Simplify the BN based on $M$
      \LineComment{{\color{teal!70}Run the IBIA framework to partition the BN}}
      \State $L_{CTF}\gets$ IBIA.run($BN,~mcs_p,~mcs_{im}$)  \Comment{{\color{teal!70} List containing calibrated CTFs for all partitions}}
      \LineComment{{\color{teal!70} Find partial assignment from last partition}}
      \State $M_p$ $\gets$ Traceback ($L_{CTF}$[-1]) \Comment{{\color{teal!70} Algorithm 1}}
      \State MPE.update($M_p$)
      \If{Number of partitions $==$ 1}
            \State \Return MPE
      \Else
      \State           Goto Line2
      \EndIf
      
    \end{algorithmic}
\end{algorithm}

Algorithm~\ref{alg:inferMPE} shows the overall procedure. All known states are contained in list $M$, initialized to evidence variables. The BN structure is simplified by removing all outgoing edges of variables with known states, and the CPDs are reduced over the known states. Next, the IBIA framework is run to partition the reduced BN based on clique size constraints $mcs_p$ and $mcs_{im}$. The framework returns the list of calibrated CTFs for all partitions. We find a partial assignment ($M_p$) over variables in the last partition using the traceback method described in Algorithm~\ref{alg:traceback}. The overall MPE assignment is updated by adding the decoded states in $M_p$. If multiple partitions are obtained, the assignment from the last partition is incomplete. Therefore, the partial assignment is added to $M$, and the process is repeated until the complete assignment is obtained. 

Our method performs a single search operation over the set of possible assignments for all variables in the BN. Due to the approximations involved while partitioning using IBIA, the beliefs in the last partition are approximate. Therefore, the partial assignment obtained in the last partition is not guaranteed to be accurate. 
The complexity can be analyzed as follows.
Let $I$ be the number of iterations required to get an assignment of all variables, and $P$  be the maximum number of partitions obtained over all iterations. The time complexity of incremental clique tree formation is polynomial in the number of variables present in the cliques that are impacted by the addition of new nodes. For each partition, both calibration using max-product BP and max-marginalization of beliefs are exponential in the max-clique size $mcs_p$. Similarly, finding the MPE assignment over variables in the last partition involves finding the argmax over individual cliques, which is also exponential in the max-clique size. Therefore, the overall time complexity is upper bounded by $O(I\cdot P \cdot 2^{mcs_p})$.

\section{Results}\label{sec:results}
All experiments presented here were carried out on 3.7-GHz Intel i7-8700 Linux system with 64-GB memory. A time limit of one hour was set for each simulation. For comparison, we use methods Weighted Mini-Bucket (WMB)~\citep{Liu2011}, Join Graph Linear Programming (JGLP)~\citep{Ihler2012} from the Merlin library and the AND/OR Branch-and-Bound (AOBB)~\citep{Dechter2007, Marinescu2009} method from the Daoopt library. The source codes for both libraries written in C++ are publicly available~\citep{Merlin,Daoopt}. Our code has been implemented in Python3 with NetworkX and NumPy libraries. All methods reduce the BN based on inherent determinism and local structure. For a fair comparison, we also simplify the BN based on known evidence states and determinism present in CPDs. We evaluate our approach on three datasets, namely, $Pedigree$, $BN$, $Grid\mbox{-}BN$~\cite{IhlerURL, MpeURL} included in the Probabilistic Inference Challenge~\citep{PIC} and several UAI approximate inference challenges. We evaluated all instances available in each dataset and ignored ones that had a single partition. We use solutions provided by ~\cite{IhlerURL} as the exact MPE solution ($X^{*}_{Exact}$).



We denote the MPE assignment obtained using an approximate inference algorithm as $X^{*}_{Alg}$. The probability of a complete assignment is simply the product of the CPDs reduced over the given state. For instances where exact solutions are available, we measure the error in the estimated max-marginals and the probability of the MPE assignment and denote it as,
\begin{align*}
    \Delta_{MaxMarg} &=\log P(MaxMarg_{Alg})-\log P(X^{*}_{Exact})\\
    \Delta_{MPE} &= \log P(X^{*}_{Alg})-\log P(X^{*}_{Exact})
\end{align*}
The runtime parameters used for various codes were as follows.
We set $mcs_p=20$ and $mcs_{im}=15$ in our code, as a trade-off between run time and accuracy. However, $mcs_{im}$ is soft constraint and its value is decremented until the network can be partitioned successfully. For WMB, JGLP and AOBB, we tried an $ibound$ of 10 and 20. Solutions for larger number of testcases were obtained for WMB with $ibound=20$ and for AOBB with $ibound=10$. Therefore, we used $ibound=20$ for WMB and JGLP. and $ibound=10$ for AOBB. 
For AOBB, we used the Mini-Bucket Elimination with Moment Matching (MBE-MM) heuristic~\citep{Flerova2011} to estimate bounds and enabled the use of the stochastic greedy ordering scheme.
\begin{table}[tb]\centering
\caption{\footnotesize Comparison of errors in log probabilities of MPE assignments ($\Delta_{MPE}$) obtained using different methods.  $P$ denotes the maximum number of partitions and $I$ denotes the number of iterations required to obtain a complete assignment with IBIA. Entries are marked as `N' if either no solution is obtained in 1hr or the probability of the solution obtained is zero.}\label{tab:deltaMap.exact}
\scriptsize
\setlength\tabcolsep{2pt}
\begin{tabular}{lrrrrrrrrrrr}\toprule
\multirow{2}{*}{} &\multirow{2}{*}{\bf Network} &\multirow{2}{*}{\bf $I$} &\multirow{2}{*}{\bf $P$} &\multicolumn{4}{c}{$\mathbf{\Delta_{MPE}}$} &\multicolumn{4}{c}{\textbf{Runtime (s)}} \\\cmidrule{5-12}
&& & &IBIA &AOBB &WMB &JGLP &AOBB &WMB &JGLP &IBIA \\\midrule
\multirow{12}{*}{\bf \rom{1}} &BN\_127 & 2 &19  &0 &0 &-53.65 &0 &0.09 &242 &216 &13 \\
&BN\_133 & 3&17 &0 &0 &-40.02 &0 &0.06 &247 &231 &14 \\
&BN\_131 & 3&17 &0 &0 &-41.02 &0 &0.04 &220 &205 &13 \\
&BN\_134 & 3&15 &0 &0 &-26.17 &0 &0.1 &245 &232 &12 \\
&BN\_130 & 2&18 &0 &0 &-61.82 &-0.17 &1 &239 &243 &10 \\
&BN\_126 & 3&19 &0 &0 &-28.40 &-19.48 &1 &208 &193 &13 \\
&BN\_129 & 3&15 &0 &0 &-39.66 &-6.40 &0.2 &196 &216 &10 \\
&BN\_132 & 2&13 &-0.31 &0 &-39.37 &-4.65 &37 &237 &228 &9 \\
&BN\_15 & 2&6 &-0.09 &0 &-2.87 &-0.66 &0.5 &106 &72 &1 \\
&pedigree30 &4&6 &0 &0 &-2.71 &0 &7 &260 &423 &14 \\
&pedigree44 &3&8 &0 &0 &-7.63 &-0.97 &492 &816 &1318 &6 \\
&pedigree13 &4&11 &-1.27 &0 &-11.09 &-1.68 &3600 &1215 &571 &10 \\
\midrule
\multirow{10}{*}{\bf \rom{2}} &BN\_60 & 2&6 &0 &0 &N&N&41 &391 &271 &3 \\
&BN\_74 &4 &10 &0 &0 &N&N&798 &50 &270 &9 \\
&pedigree23  &2&3 &0 &0 &N&N&0.1 &25 &3600 &1 \\
&pedigree42 & 4&7 &-0.09 &0 &N&N&273 &63 &731 &3 \\
&pedigree7 &5&10 &-0.37 &0 &N&-2.68 &3600 &3600 &2467 &10 \\
&pedigree9 &4&9 &-0.95 &0 &N&-1.03 &379 &808 &513 &10 \\
&pedigree34 &4 &10&-1.18 &0 &N&N&3600 &1505 &3601 &8 \\
&pedigree41 &3&9 &-1.06 &0 &N&N&3600 &47 &3601 &8 \\
&pedigree33 &2&3 &-1.91 &0 &N&-0.80 &6 &2276 &460 &3 \\
&pedigree31 &5 &13&-2.25 &-0.45 &N&-3.34 &3600 &3601 &1743 &12 \\
&pedigree40  &4 &11&-1.88 &-7.89 &N&N&3600 &90 &196 &12 \\
&pedigree51 &5 &10 &0 &-4.68 &N&-2.30 &3600 &1866 &750 &11 \\
\bottomrule
\end{tabular}
\end{table}

\begin{table}[!htbp]\centering
\caption{\footnotesize Comparison of the logarithm of MPE assignment probabilities obtained with different methods. $P$ denotes the maximum number of partitions and $I$ denotes the number of iterations required to obtain a complete assignment with IBIA. Entries are marked as `N' if either no solution is obtained in 1hr or the probability of the solution obtained is zero.}\label{tab:map.noexact}
\scriptsize
\setlength\tabcolsep{2pt}
\begin{tabular}{lrrrrrrrrrrr}\toprule
\multirow{2}{*}{} &\multirow{2}{*}{\bf Network}&\multirow{2}{*}{\bf $I$} &\multirow{2}{*}{\bf $P$} &\multicolumn{4}{c}{$\log P(X^{*}_{Alg})$} &\multicolumn{4}{c}{\textbf{Runtime (s)}} \\\cmidrule{5-12}
&& & &IBIA &AOBB &WMB &JGLP &AOBB &WMB &JGLP &IBIA \\\midrule
\multirow{9}{*}{\bf \rom{1}}&75-21-5&2 &3 &-13.28 &-13.28 &N &-13.37 &21 &229 &140 &2 \\
&90-20-5&2 &2 &-5.70 &-5.70 &N &-6.62 &0.3 &186 &122 &1  \\
&75-22-5&2 &5 &-15.61 &-15.61 &N &-16.46 &6 &227 &168 &1 \\
&75-23-5&3 &6 &-15.43 &-15.43 &N &-16.40 &82 &299 &188 &2 \\
&75-24-5&3 &5 &-16.08 &-16.08 &N &-16.87 &37 &331 &211 &2 \\
&75-26-5&2 &6 &-21.89 &-21.89 &N&-23.06& 3600 &417 &270 & 4 \\
&75-25-5&2 &6 &-20.84 &-20.84 &N &N & 520 &327 &235 & 3 \\
&90-30-5&3 &5 &-13.12 &-13.12 &N &N &1428 &456 &378 &4  \\
&90-34-5&3 &3 &-13.29 &-13.29 &N &N &898 &645 &489 &4 \\
\midrule
\multirow{8}{*}{\bf \rom{2}}&BN\_72&17 &22 &-193.89 &N &N &N  &3600 &3600 &2570 &141\\
&BN\_70&8 &21 &-126.62 &-136.58 &N &N  &3600 &2606 & 1306 &138 \\
&BN\_71&12 &23 &-157.85 &N &N &N & 3600 &1785 &471 & 129 \\
&BN\_73&13 &21 &-169.45 &-180.98 &N &N &3600 &3600 &2079 &140 \\
&90-38-5&5 &11 &-19.87 &N &N &N &3600 &906 &581  &13\\
&90-42-5&6 &9 &-20.94 &N &N &N &3600 &1101 &751 &16 \\
&90-46-5&3  &10 &-28.32 &N &N &N &3600 &1324 &923 &13  \\
&90-50-5&5 &13 &-29.16 &N &N &N  &3600 &1547 &1055 &24 \\
\bottomrule
\end{tabular}
\end{table}
Table~\ref{tab:deltaMap.exact} shows a detailed comparison of the error in MPE assignments ($\Delta_{MPE}$) and the required runtime for different inference methods for a subset of benchmarks. The results for others are similar. For each testcase, the table also shows the maximum number of partitions and the number of iterations required to obtain a complete MPE assignment with IBIA. We check for accuracy of log probabilities upto two decimal places and report errors $<0.01$ as 0. Testcases for which solutions obtained by all methods were close to the optimal solution are not tabulated here in the interest of space. We divide testcases into two categories based on the results. For benchmarks in Category~\rom{1}, all methods achieve a non-zero probability solution. IBIA gives close to optimal solutions for most of the benchmarks in this category. Category~\rom{2} shows testcases where IBIA and AOBB give solutions, but either WMB or JGLP fails. IBIA achieves good estimates in most testcases. For testcases $BN\_74,~pedigree7,~pedigree51$, while IBIA achieves close to the optimal solution in $<11$ seconds, AOBB  performs repeated search operations until the timeout constraint is reached. While the error obtained with IBIA is $>1$ in five testcases, in four out of five of these testcases, AOBB also performs repeated search until timeout. IBIA performs better than JGLP in all cases except $pedigree33$. For testcases $pedigree40$ and $pedigree51$, IBIA provides a higher probability solution than other methods.

In Table~\ref{tab:map.noexact}, we compare the logarithm of the MPE probabilities obtained with the different methods for benchmarks where the exact solution is not available. We ignore cases where IBIA, AOBB, and WMB/JGLP give a common solution while tabulation. For testcases in the first category in~Table~\ref{tab:map.noexact}, IBIA gives the same solution as AOBB in $<5$ seconds. The estimate obtained in all these cases is better than both WMB and JGLP. For networks in the second category, IBIA performs better than all other methods. While AOBB achieves some non-zero solution in testcases $BN\_70$ and $BN\_73$, the solution obtained with IBIA has a higher probability than the AOBB solution, with the difference being around 10. In all other cases, IBIA is able to find a possible MPE assignment with non-zero probability while all other methods obtain a zero-probability solution within the set time limit.

Since IBIA is written in Python and available tools are written in C++, the runtimes reported here are not directly comparable. That said, the runtime for IBIA is an order of magnitude lower than JGLP and WMB in most cases. They are also better than AOBB in many cases. This is expected since IBIA does not search for solutions. A single partial assignment is obtained in each iteration.

Table~\ref{tab:testcaseCount} summarizes the total number of instances that require multiple partitions and the number of instances that provide non-zero probability assignments within 1hr for different benchmarks. Out of 117 networks, IBIA provides solutions in 100 testcases which is better than JGLP and WMB, but slightly lower than 108 cases solved by AOBB. 

The approximate CT obtained after local max-marginalization depends on the choice of variables that are removed to reduce the clique size. For the results in Tables \ref{tab:deltaMap.exact} and\ref{tab:map.noexact}, we prioritize variables present in the least number of cliques. A random assignment is chosen in case of ties. The CTF obtained after approximation will be different if a different set of variables is chosen for removal. We tried three other orderings without any prioritization. Although the change in $MaxMarg$ was not much in most cases, there were significant changes in the MPE. The results for the benchmarks where an improvement was obtained are shown in Table \ref{tab:ibiaBestSol}, which has the best solution obtained. This gives rise to the possibility of searching for a solution over the approximation space. An advantage of this is that the approximation instances are independent of each other and can be run in parallel.
\begin{table}\centering
\caption{\footnotesize Number of testcases where a non-zero solution is obtained using various methods with time limit set to 1hr.}\label{tab:testcaseCount}
\scriptsize
\setlength\tabcolsep{2pt}
\begin{tabular}{ccccccc}\toprule
&\textbf{\#Inst.} &\textbf{WMB} &\textbf{JGLP} &\textbf{AOBB+} &\textbf{IBIA}\\
& & & &\textbf{MBE-MM} &
\\\midrule
\textbf{BN\_UAI} &63 &38 &42 &58 &48 \\
\textbf{Pedigree} &22 &7 &12 &22 &20 \\
\textbf{GridBN} &32 &8 &22 &28 &32 \\ \bottomrule
\textbf{Total} & 117 & 53 & 76 & 108 & 100 \\
\bottomrule
\end{tabular}
\end{table}
\begin{table}\centering
\caption{\footnotesize Comparison of errors in MPE probabilities obtained with IBIA using different orderings for selection of variables while local max-marginalization. $\Delta_{IBIA}$ is the error obtained using structure based heuristics. $\Delta_{IBIA_{Best}}$ is error of the best estimate obtained. }\label{tab:ibiaBestSol}
\scriptsize
\setlength\tabcolsep{2pt}
\begin{tabular}{lccccccc}\toprule
\bf Network & $ \mathbf{\Delta_{IBIA}}$ & $ \mathbf{\Delta_{IBIA_{Best}}}$ && \bf Network & $ \mathbf{\Delta_{IBIA}}$ & $ \mathbf{\Delta_{IBIA_{Best}}}$\\\midrule
BN\_115 &-1.51 &0 & &pedigree31 &-2.25 &0 \\
BN\_12 &-0.38 &-0.19 & &pedigree33 &-1.91 &0 \\
BN\_132 &-0.31 &0 & &pedigree34 &-1.18 &0 \\
BN\_69 &N &-0.64 & &pedigree37 &-0.46 &0 \\
BN\_90 &-16 &0 & &pedigree40 &-1.88 &-0.65 \\
pedigree13 &-1.27 &0 & &pedigree41 &-1.06 &0 \\
pedigree18 &N &0 & &pedigree42 &-0.09 &0 \\
pedigree19 &N &-0.01 & &pedigree50 &-1.25 &-1.07 \\
pedigree20 &-0.53 &-0.02 & &pedigree9 &-0.95 &0 \\
pedigree25 &-2.86 &0 & & & & \\
\bottomrule
\end{tabular}
\end{table}

\begin{figure}
    \centering
    \centering
     \includegraphics[width=0.4\textwidth]{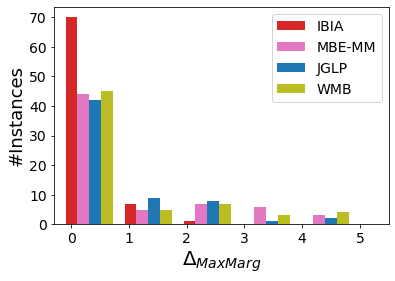}
    \caption{\footnotesize Comparison of error in max-marginal estimated using different methods for $BN$ and $Pedigree$ instances where solutions are known. We limit the range to compare the number of instances with lower errors. \#Instances: 79 Bin-size$=$1}
     \label{fig:deltaMM}
\end{figure}

To evaluate the quality of the estimate of max-marginal obtained with our method, we compare it with the final upper bounds achieved with JGLP, WMB, and MBE-MM. Figure~\ref{fig:deltaMM} shows the histogram of error in estimates obtained with different methods. We observe that the error with IBIA is within $\pm1$ in 70 instances out of 79 cases. However, the upper bound with all other methods is in this range in slightly more than 40 testcases. The average error in the estimate of max-marginal obtained with IBIA is 0.27 as opposed to 1.12, 1.23, 2.17 for the upper bounds estimated using existing metrics WMB, JGLP, and MBE-MM.
Though the estimate of max-marginal obtained with IBIA is not necessarily an upper bound, it provides good estimates when compared to the existing heuristics. Therefore, this method can be used for quick evaluation of different partial assignments in search algorithms like Branch and Bound.

\section{Conclusions}
We propose a single-shot algorithm for MPE inference based on an Incremental Build-Infer-Approximate paradigm. 
It is an iterative method in which the
subset of variables for which an assignment is obtained
is guaranteed to increase in every iteration. 
In majority of cases, the number of iterations required is $<5$.
Even though it is a single shot algorithm, it gives valid assignments in 100 out of 117 testcases. In majority of the testcases, the accuracy is comparable to the branch and bound based technique AOBB. Overall the runtimes are competitive, with significantly lower runtimes than AOBB in several cases. We also show that our method achieves better estimates of the max-marginals when compared to the upper bounds obtained with several mini-bucket methods. Therefore, this approach can possibly be used in search algorithms like Branch and Bound for quick evaluation of different partial assignments.
 We show some initial experiments to improve the accuracy of the approach by using different heuristics for approximation.
 
We obtain zero probability solutions in 14 out of a total
of 117 benchmarks, calling for a more systematic search over the approximation space. In many of these cases, there were multiple partial assignments, and we chose one of them randomly. A search over these could also possibly give better solutions.

\bibliography{ref}

\begin{thebibliography}{35}
\providecommand{\natexlab}[1]{#1}
\providecommand{\url}[1]{\texttt{#1}}
\expandafter\ifx\csname urlstyle\endcsname\relax
\providecommand{\doi}[1]{doi: #1}\else
\providecommand{\doi}{doi: \begingroup \urlstyle{rm}\Url}\fi

\bibitem[Bathla and Vasudevan(2021)]{Bathla2021}
Shivani Bathla and Vinita Vasudevan.
\newblock A memory constrained approximate bayesian inference approach using
incremental construction of clique trees.
\newblock TechRxiv. https://doi.org/10.36227/techrxiv.14938275.v1, Jul
2021.

\bibitem[Bathla and Vasudevan(2022)]{JAIR}
Shivani Bathla and Vinita Vasudevan.
\newblock {IBIA}: Bayesian inference via incremental build-infer-approximate
operations on clique trees.
\newblock \emph{arXiv preprint arXiv:2202.12003}, 2022.

\bibitem[Bauer et~al.(2019)Bauer, Nakajima, G{\"o}rnitz, and
M{\"u}ller]{Bauer2019}
Alexander Bauer, Shinichi Nakajima, Nico G{\"o}rnitz, and Klaus-Robert
M{\"u}ller.
\newblock Partial optimality of dual decomposition for map inference in
pairwise mrfs.
\newblock In \emph{The 22nd International Conference on Artificial Intelligence
and Statistics}, pages 1696--1703, 2019.

\bibitem[Dechter(2006)]{MpeURL}
Rina Dechter.
\newblock Uai model files and solutions.
\newblock
https://www.ics.uci.edu/\textasciitilde dechter/softwares/benchmarks/\\
Mpe\_Problme\_Sets/,
2006.
\newblock Accessed: 2021-10-15.

\bibitem[Dechter and Mateescu(2007)]{Dechter2007}
Rina Dechter and Robert Mateescu.
\newblock And/or search spaces for graphical models.
\newblock \emph{Artificial intelligence}, 171\penalty0 (2-3):\penalty0 73--106,
2007.

\bibitem[Elidan(2011)]{PIC}
Gal Elidan.
\newblock The probabilistic inference challenge (pic2011).
\newblock https://www.cs.huji.ac.il/project/PASCAL/, 2011.

\bibitem[Flerova et~al.(2011)Flerova, Ihler, Dechter, and Otten]{Flerova2011}
Natalia Flerova, Alexander Ihler, Rina Dechter, and Lars Otten.
\newblock Mini-bucket elimination with moment matching.
\newblock In \emph{NIPS Workshop DISCML}, 2011.

\bibitem[Flores et~al.(2002)Flores, G{\'{a}}mez, and Olesen]{Flores2002}
M~Julia Flores, Jos{\'{e}}~A G{\'{a}}mez, and Kristian~G Olesen.
\newblock {Incremental Compilation of Bayesian Networks}.
\newblock In \emph{Proceedings of the Nineteenth Conference on Uncertainty in
Artificial Intelligence}, pages 233--240, 2002.

\bibitem[Globerson and Jaakkola(2007)]{Globerson2007}
Amir Globerson and Tommi Jaakkola.
\newblock Fixing max-product: Convergent message passing algorithms for map
lp-relaxations.
\newblock \emph{Advances in neural information processing systems}, 20, 2007.

\bibitem[Hazan and Shashua(2010)]{Hazan2010}
Tamir Hazan and Amnon Shashua.
\newblock Norm-product belief propagation: Primal-dual message-passing for
approximate inference.
\newblock \emph{IEEE Transactions on Information Theory}, 56\penalty0
(12):\penalty0 6294--6316, 2010.

\bibitem[Ihler(2006)]{IhlerURL}
Alexander Ihler.
\newblock Uai model files and solutions.
\newblock http://sli.ics.uci.edu/\textasciitilde ihler/uai\-data/, 2006.
\newblock Accessed: 2021-10-15.

\bibitem[Ihler et~al.(2012)Ihler, Flerova, Dechter, and Otten]{Ihler2012}
Alexander~T Ihler, Natalia Flerova, Rina Dechter, and Lars Otten.
\newblock Join-graph based cost-shifting schemes.
\newblock \emph{Proceedings of the Twenty-Eighth Conference on Uncertainty in
Artificial Intelligence}, 2012.

\bibitem[Jojic et~al.(2010)Jojic, Gould, and Koller]{Jojic2010}
Vladimir Jojic, Stephen Gould, and Daphne Koller.
\newblock Accelerated dual decomposition for map inference.
\newblock In \emph{Proceedings of the 27th International Conference on
International Conference on Machine Learning}, page 503–510, 2010.

\bibitem[Koller and Friedman(2009)]{Koller2009}
Daphne Koller and Nir Friedman.
\newblock \emph{Probabilistic graphical models: principles and techniques}.
\newblock MIT press, 2009.

\bibitem[Kolmogorov(2005)]{Kolmogorov2005}
Vladimir Kolmogorov.
\newblock Convergent tree-reweighted message passing for energy minimization.
\newblock In \emph{International Workshop on Artificial Intelligence and
Statistics}, pages 182--189, 2005.

\bibitem[Komodakis et~al.(2007)Komodakis, Paragios, and
Tziritas]{Komodakis2007}
Nikos Komodakis, Nikos Paragios, and Georgios Tziritas.
\newblock Mrf optimization via dual decomposition: Message-passing revisited.
\newblock In \emph{2007 IEEE 11th International Conference on Computer Vision},
pages 1--8, 2007.

\bibitem[Liu and Ihler(2011)]{Liu2011}
Qiang Liu and Alexander Ihler.
\newblock Bounding the partition function using holder's inequality.
\newblock In \emph{Proceedings of the 28th International Conference on Machine
Learning}, pages 849--856, 2011.

\bibitem[Marinescu(2016)]{Merlin}
Radu Marinescu.
\newblock Merlin.
\newblock https://github.com/radum2275/merlin/, 2016.
\newblock Accessed: 2021-10-15.

\bibitem[Marinescu and Dechter(2005)]{Marinescu2005}
Radu Marinescu and Rina Dechter.
\newblock And/or branch-and-bound for graphical models.
\newblock In \emph{Proceedings of the 19th International Joint Conference on
Artificial Intelligence}, pages 224--229, 2005.

\bibitem[Marinescu and Dechter(2007)]{Marinescu2007}
Radu Marinescu and Rina Dechter.
\newblock Best-first and/or search for most probable explanations.
\newblock In \emph{Proceedings of the Twenty-Third Conference on Uncertainty in
Artificial Intelligence}, pages 259--266, 2007.

\bibitem[Marinescu and Dechter(2009)]{Marinescu2009}
Radu Marinescu and Rina Dechter.
\newblock Memory intensive and/or search for combinatorial optimization in
graphical models.
\newblock \emph{Artificial Intelligence}, 173\penalty0 (16-17):\penalty0
1492--1524, 2009.

\bibitem[Marinescu et~al.(2014)Marinescu, Dechter, and Ihler]{Marinescu2014}
Radu Marinescu, Rina Dechter, and Alexander~T Ihler.
\newblock And/or search for marginal map.
\newblock In \emph{Proceedings of the Thirtieth Conference on Uncertainty in
Artificial Intelligence}, pages 563--572, 2014.

\bibitem[Marinescu et~al.(2018)Marinescu, Lee, Dechter, and
Ihler]{Marinescu2018}
Radu Marinescu, Junkyu Lee, Rina Dechter, and Alexander Ihler.
\newblock And/or search for marginal map.
\newblock \emph{Journal of Artificial Intelligence Research}, 63:\penalty0
875--921, 2018.

\bibitem[Marinescu et~al.(2020)Marinescu, Kishimoto, and Botea]{Marinescu2020}
Radu Marinescu, Akihiro Kishimoto, and Adi Botea.
\newblock Parallel and/or search for marginal map.
\newblock In \emph{Proceedings of the AAAI Conference on Artificial
Intelligence}, pages 10226--10234, 2020.

\bibitem[Martins et~al.(2015)Martins, Figueiredo, Aguiar, Smith, and
Xing]{Martins2015}
Andr{\'e}~FT Martins, M{\'a}rio~AT Figueiredo, Pedro~MQ Aguiar, Noah~A Smith,
and Eric~P Xing.
\newblock Ad3: Alternating directions dual decomposition for map inference in
graphical models.
\newblock \emph{The Journal of Machine Learning Research}, 16\penalty0
(1):\penalty0 495--545, 2015.

\bibitem[Otten(2010)]{Daoopt}
Lars Otten.
\newblock Daoopt: Distributed and/or optimization.
\newblock https://github.com/lotten/daoopt/, 2010.
\newblock Accessed: 2022-02-10.

\bibitem[Otten and Dechter(2012)]{Otten2012}
Lars Otten and Rina Dechter.
\newblock Anytime and/or depth-first search for combinatorial optimization.
\newblock \emph{Ai Communications}, 25\penalty0 (3):\penalty0 211--227, 2012.

\bibitem[Park and Darwiche(2004)]{Park2004}
James~D Park and Adnan Darwiche.
\newblock Complexity results and approximation strategies for map explanations.
\newblock \emph{Journal of Artificial Intelligence Research}, 21:\penalty0
101--133, 2004.

\bibitem[Ping et~al.(2015)Ping, Liu, and Ihler]{Ping2015}
Wei Ping, Qiang Liu, and Alexander~T Ihler.
\newblock Decomposition bounds for marginal map.
\newblock \emph{Advances in neural information processing systems}, 28, 2015.

\bibitem[Shimony and Charniak(1990)]{Shimony1990}
Solomon~Eyal Shimony and Eugene Charniak.
\newblock A new algorithm for finding map assignments to belief networks.
\newblock In \emph{Proceedings of the Sixth Annual Conference on Uncertainty in
Artificial Intelligence}, pages 185--196, 1990.

\bibitem[Sontag et~al.(2008)Sontag, Meltzer, Globerson, Jaakkola, and
Weiss]{Sontag2008}
David Sontag, Talya Meltzer, Amir Globerson, Tommi Jaakkola, and Yair Weiss.
\newblock Tightening lp relaxations for map using message passing.
\newblock In \emph{Proceedings of the Twenty-Fourth Conference on Uncertainty
in Artificial Intelligence}, pages 503--510, 2008.

\bibitem[Sontag et~al.(2011)Sontag, Globerson, and Jaakkola]{Sontag2011}
David Sontag, Amir Globerson, and Tommi Jaakkola.
\newblock Introduction to dual composition for inference.
\newblock In \emph{Optimization for Machine Learning}. MIT Press, 2011.

\bibitem[Sontag et~al.(2012)Sontag, Choe, and Li]{Sontag2012}
David Sontag, Do~Kook Choe, and Yitao Li.
\newblock Efficiently searching for frustrated cycles in map inference.
\newblock In \emph{Proceedings of the Twenty-Eighth Conference on Uncertainty
in Artificial Intelligence}, pages 795--804, 2012.

\bibitem[Wainwright et~al.(2005)Wainwright, Jaakkola, and
Willsky]{Wainwright2005}
Martin~J Wainwright, Tommi~S Jaakkola, and Alan~S Willsky.
\newblock Map estimation via agreement on trees: message-passing and linear
programming.
\newblock \emph{IEEE transactions on information theory}, 51\penalty0
(11):\penalty0 3697--3717, 2005.

\bibitem[Werner(2007)]{Werner2007}
Tomas Werner.
\newblock A linear programming approach to max-sum problem: A review.
\newblock \emph{IEEE transactions on pattern analysis and machine
intelligence}, 29\penalty0 (7):\penalty0 1165--1179, 2007.

\end{thebibliography}
\section{Supplementary Material}\label{sec:sup}

%

In the following, we use the terms $CTF_{in}$ and $CTF_a$ to denote the input and output CTF of Algorithm~\ref{alg:ApproximateCTF}. The line numbers indicated here refer to line numbers of Algorithm~\ref{alg:ApproximateCTF}.\\

\textbf{Proof for Proposition~\ref{pr:approx1}}
$CTF_a$ is initialized as the subgraph of $CTF_{in}$ which is valid and max-calibrated (line 2). Each CT $\in CTF_a$ obtained after the subsequent steps is valid because of the following reasons.
\begin{itemize}
	\item It contains only maximal cliques since
	any non-maximal cliques generated while simplification of $CTF_a$ are removed (lines 6,22).
	\item It contains disjoint trees. This is because neither of exact and local max-marginalization introduce any loops. 
	In the exact max-marginalization step, neighbors of the collapsed and non-maximal cliques are reconnected to $CTF_a$, which means that the tree structure of the CT is preserved. 
	In the local max-marginalization step, variables are removed only if no sepset size becomes zero, ensuring that a connected CT remains a connected tree (lines 15-18). 
	\item It satisfies the running intersection property (RIP)~\citep{Koller2009}.
	This is because
	\begin{itemize}
		\item During exact max-marginalization, sepsets in the $CTF_{a}$ are preserved while re-connecting the neighbors of the collapsed cliques to the new clique, thereby ensuring that RIP is satisfied.
		Similarly, since neighbors of non-maximal cliques which are removed are connected to the containing cliques, sepsets are preserved.		
		\item During local max-marginalization, variables are retained in a single connected component of $CTF_a$ (line 14).
		Therefore, the subgraph of $CTF_a$ over any net is connected. 				
	\end{itemize}
      \item  The re-assignment of factors results in a valid distribution. \\
\textit{Proof:} After exact max-marginalization, all the clique and sepset beliefs are preserved. Therefore, the resultant clique is also calibrated.
Let $C_i$ and $C_j$ be two adjacent cliques in $CTF_{in}$ with sepset $S_{i,j}$. After local max-marginalization of a node $n$, we get the corresponding cliques $C_i', C_j' \in CTF_a$, with sepset $S_{i,j}'$. Since the result is invariant with respect to the order in which variables are maximized over, we have
  \begin{align*}
           \max_{{C_i}'\setminus {S_{i,j}}'}\beta(C_i') &= \max_{{C_i}'\setminus {S_{i,j}}'} \max_{n.states}\beta(C_i) \\
           &= \max_{n.states}\max_{{C_i}'\setminus {S_{i,j}}'}\beta(C_i) \\
           &=  \max_{n.states}\mu(S_{i,j})
           = \mu(S_{i,j}')
   \end{align*}
    where  $n.states$ denotes the states of $n$. Similarly, $\max_{{C_j}'\setminus {S_{i,j}}'}\beta(C_j') = \mu(S_{i,j}')$. Since this is true for all pairs of adjacent cliques, $CTF_a$ is also  max-calibrated. Therefore, since re-parameterization is performed using calibrated clique and sepset beliefs, it results in a valid distribution.
    \end{itemize}

\textbf{Proof for Proposition~\ref{pr3}}
 $CTF_a$ is obtained from $CTF_{in}$, which is calibrated. Therefore, all cliques in a CT have the same maximum belief. After exact max-marginalization, the clique beliefs in $CTF_a$ are identical to the beliefs in $CTF_{in}$,  since the step involves collapsing all containing cliques before max-marginalization of beliefs. Both exact and local max-marginalizations involve maximizing clique beliefs over all states of variable. Thus, the the maximum belief  and the within clique beliefs remain unaltered. 


\end{document}